\documentclass[conference]{IEEEtran}
\IEEEoverridecommandlockouts

\usepackage{cite}
\usepackage{amsmath,amssymb,amsfonts}
\usepackage{algorithmic}
\usepackage{graphicx}
\usepackage{textcomp}
\usepackage{xcolor}
\usepackage{comment}
\usepackage{multirow} 
\usepackage{caption} 
\usepackage{hyperref}

\def\BibTeX{{\rm B\kern-.05em{\sc i\kern-.025em b}\kern-.08em
    T\kern-.1667em\lower.7ex\hbox{E}\kern-.125emX}}
\begin{document}

\title{Interpretability-Aware Vision Transformer}

\author{
Yao Qiang, Chengyin Li, Hui Zhu, Prashant Khanduri, Dongxiao Zhu \\
Computer Science Department, Wayne State University, Detroit, USA\\
Email: \{yao, cyli, hq2197, khanduri.prashant, dzhu\}@wayne.edu
}

\maketitle

\begin{abstract}

    Vision Transformers (ViTs) have become prominent models for solving various vision tasks. However, the interpretability of ViTs has not kept pace with their promising performance. While there has been a surge of interest in developing {\it post hoc} solutions to explain ViTs' outputs, these methods do not generalize to different downstream tasks and various transformer architectures. Furthermore, if ViTs are not properly trained with the given data and do not prioritize the region of interest, the {\it post hoc} methods become less effective. To overcome this limitation, we introduce a novel training procedure that inherently enhances ViT's interpretability. Our interpretability-aware ViT (IA-ViT) draws inspiration from a fresh insight: both the class patch and image patches consistently generate predicted distributions and attention maps. IA-ViT is composed of a feature extractor, a predictor, and an interpreter, which are trained jointly with an interpretability-aware training objective. Consequently, the interpreter simulates the behavior of the predictor and provides a faithful explanation through its single-head self-attention mechanism. Our comprehensive experimental results demonstrate the effectiveness of IA-ViT in several image classification tasks, with both qualitative and quantitative evaluations of model performance and interpretability. Our
    code is available at: https://github.com/qiangyao1988/IA-ViT.
\end{abstract}


\section{Introduction}
\label{introduction}

The Transformer architecture \cite{vaswani2017attention}, originally designed for natural language processing (NLP) tasks \cite{devlin2018bert}, has recently found application in computer vision (CV) tasks with the emergence of Vision Transformer (ViT) \cite{dosovitskiy2020image}. ViT utilizes the multi-head self-attention (MSA) mechanism as its foundation, enabling it to proficiently capture long-range dependencies among pixels or patches within images. As a result, ViTs have demonstrated superior performance over state-of-the-art convolutional neural networks (CNNs) in numerous CV tasks, including but not limited to image classification \cite{liu2021dimbert,liu2021swin,touvron2021training,yuan2021tokens,xu2022privacy}, object detection \cite{carion2020end,chu2021twins,wang2022bevt,wang2021pyramid}, action recognition \cite{liu2022video,zhang2021vidtr}, and medical imaging segmentation \cite{li2022focalunetr,li2023auto}.

Since ViTs are extensively employed in high-stakes decision-making fields like healthcare \cite{stiglic2020interpretability} and autonomous driving \cite{kim2017interpretable}, there exists a significant demand for gaining insights into their decision-making process. Nonetheless, ViTs continue to function as black-box models, lacking transparency and explanations for both their training process and predictions. Explainable AI (XAI) has arisen as a specialized field within AI, with the goal of ensuring that end users intuitively understand and trust the models' outputs by providing explanations for their behaviors \cite{samek2019explainable,arrieta2020explainable,qiang2022counterfactual}. 

XAI encompasses numerous research directions. One strand focuses on  {\it post hoc} explanation techniques, which aim to obtain explanations by approximating a pre-trained model and its predictions \cite{ribeiro2016should,zhou2016learning,lundberg2017unified,shrikumar2017learning,sundararajan2017axiomatic,petsiuk2018rise,pan2021explaining,qiang2024fairness,li2023negative}. Although there has been an increasing interest in developing {\it post hoc} solutions for Transformers, most of them either rely on the attention weights within the MSA mechanism \cite{hao2021self,abnar2020quantifying} or utilize back-propagation gradients to generate explanations \cite{chefer2021transformer,pan2021explaining,qiang2022attcat,li2023negative}. It is important to highlight that these approaches have limitations in terms of their ability to elucidate the decision-making processes of trained models and can be impacted by different input schemes \cite{alvarez2018robustness,adebayo2018sanity,kindermans2019reliability}. Conversely, a different strand of research focuses on modifying neural architectures \cite{frosst2017distilling,wu2018beyond} and/or incorporating explanations into the learning process \cite{ross2017right,ghaeini2019saliency,ismail2021improving} for better interpretability. Building explainable ViT models during training remains largely uncharted waters. Recent studies tend to modify the ViT architecture or rely on external knowledge to provide faithful explanations \cite{rigotti2021attention, kim2022vit}.

\begin{figure}[t]
    \centering
    \includegraphics[width=0.90\linewidth]{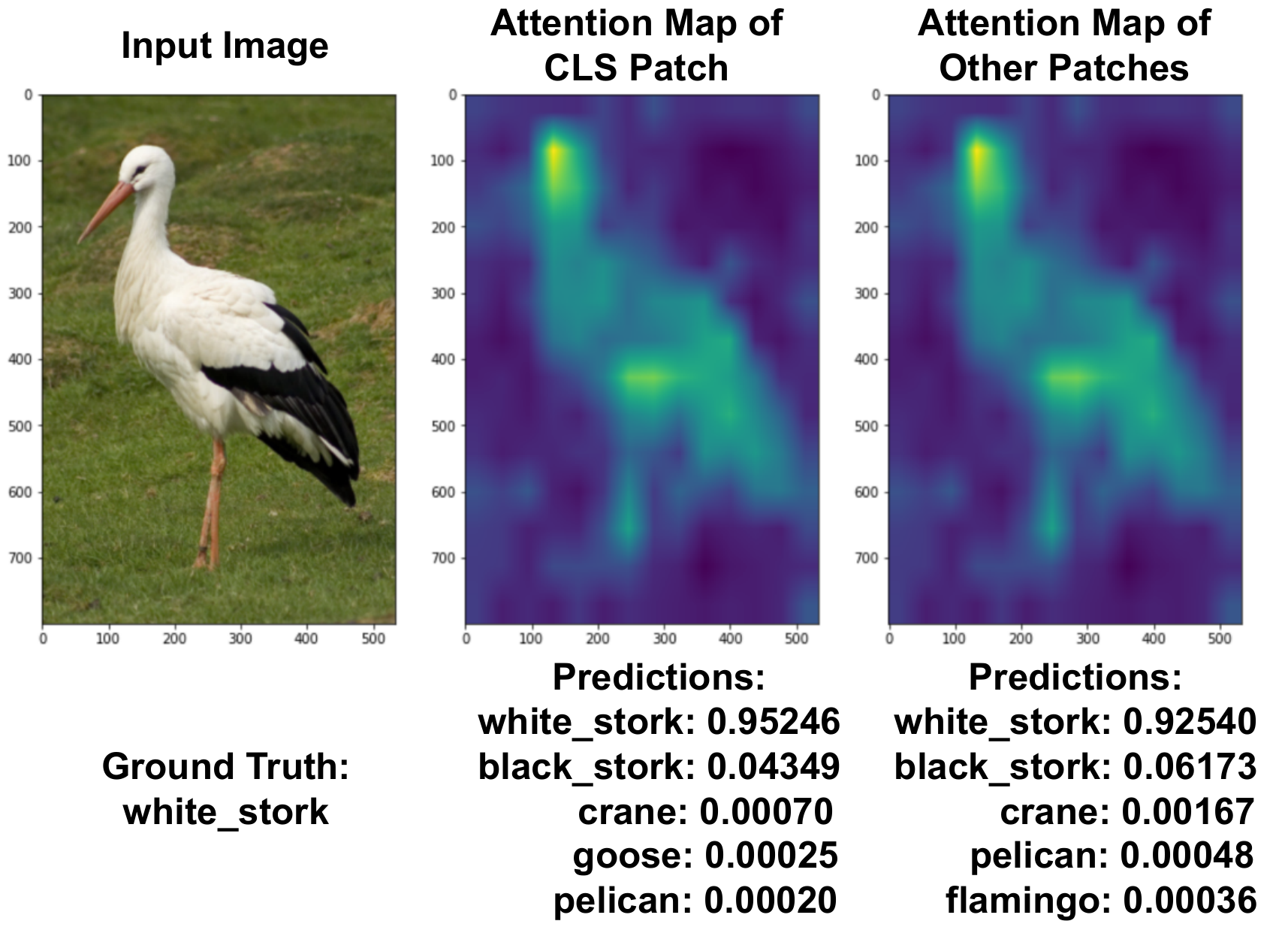}
    \vspace{-0.15cm}
    \caption{Illustration of attention maps and predictive distributions from both the CLS patch and other image patches.} 
    \vspace{-0.75cm}
    \label{fig:illustration}
\end{figure}

Among efforts to improve interpretability during training, we propose our novel interpretability-aware ViT (IA-ViT). Our inspiration comes from the observation that, in ViT models, the downstream classification tasks only utilize the embedding of the class (CLS) patch. In contrast, the feature embeddings of the image patches, which are learned using multi-layer MSA blocks, are underutilized and often neglected. However, we discover that these neglected patch embeddings also contain crucial discriminative features for classification. Both the CLS and the image patches generate uniform predictive distributions and attention maps, as illustrated in Fig.\ref{fig:illustration}. Therefore, we suggest leveraging the valuable attributes of these image patches for interpretation while utilizing the CLS patch embedding for prediction. The key is to treat interpretation and prediction as distinct but interrelated tasks. To achieve this, we introduce our IA-ViT architecture and a novel training framework to optimize both tasks simultaneously.

\begin{figure}[t]
    \centering 
    \includegraphics[width=0.95\linewidth]{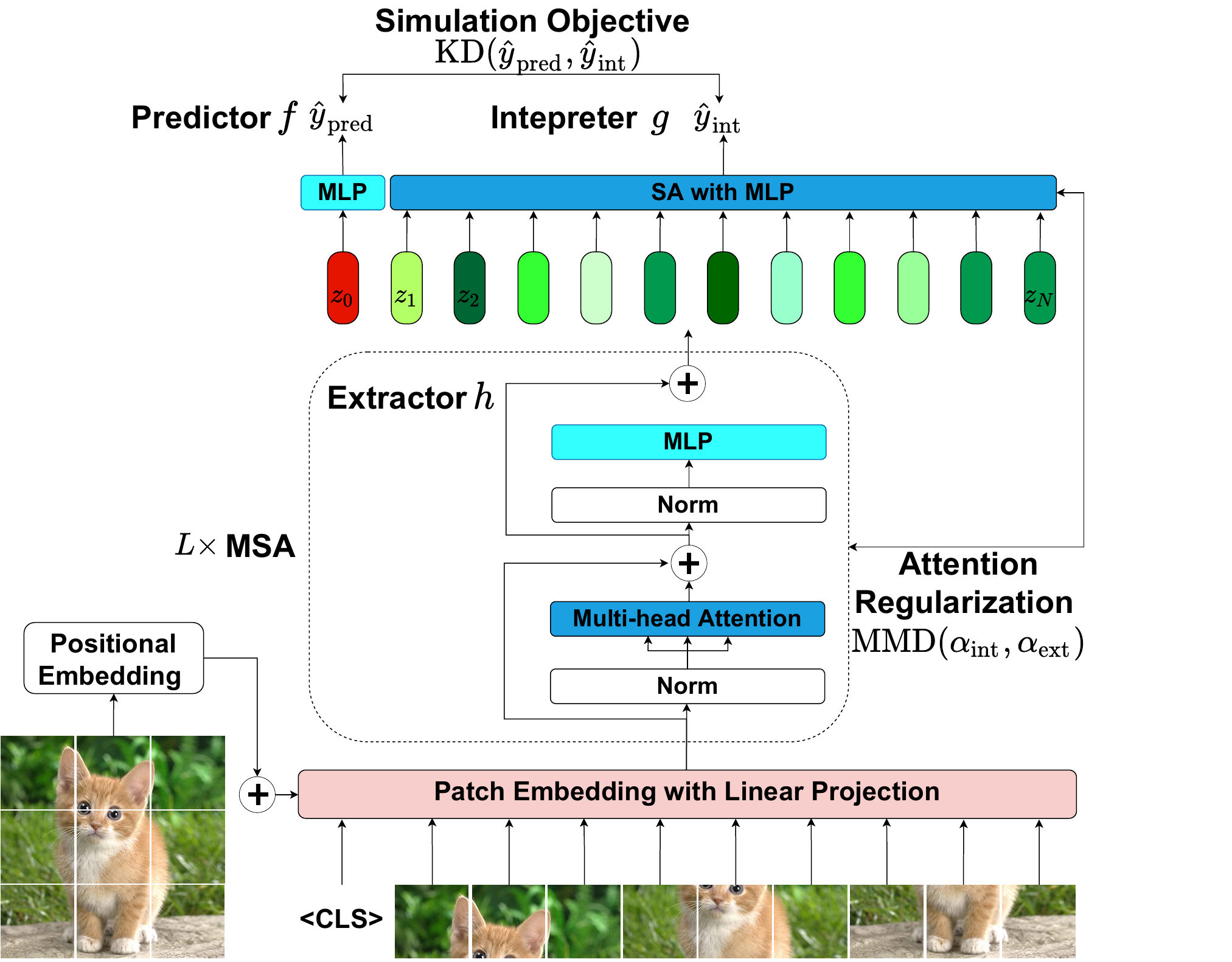}
    \caption{IA-ViT consists of three major components: feature extractor, predictor, and interpreter. Both the predictor and the interpreter generate the class prediction for this cat image. KD is applied on the two logits in the simulation objective. The attention weights in SA and MSA are aligned via MMD during the training process for better explanations.}
    \label{fig:architecture}
    \vspace{-0.7cm}
\end{figure}

As illustrated in Fig. \ref{fig:architecture}, we introduce an additional interpreter into the ViT architecture as the interpretability-aware component aside from ViT's inherent predictor. This interpreter comprises a single-head self-attention (SSA) mechanism and a linear head. SSA is employed to generate explanations through its attention weights, while the linear head maps the embeddings of image patches into the label space aiming to simulate the behavior of the predictor. In our novel \textit{Learning with Interpretation} training framework, IA-ViT employs a joint training strategy for both the predictor and the interpreter. This approach enables the interpreter to gain insights that align with the predictor’s outputs through the simulation objective, while also improving overall interpretability via attention regularization, as illustrated in Fig.\ref{fig:architecture}. Further details about the \textit{Learning with Interpretation} framework are provided in Section \ref{sec:Learning with Interpretation}. In summary, IA-ViT maintains its high expressive power while integrating an interpretability-aware training objective, offering stable and reliable explanations.

We summarize our major contributions: (1) We propose a novel ViT architecture, which leverages the feature embeddings from the image patches beside the CLS patch to provide consistent, faithful, and high-quality explanations while maintaining high predictive performance. (2) Our interpretability-aware training objective has been demonstrated effective in enhancing the interpretability of IA-ViT. (3) We conduct a comprehensive comparison of our approach with several strong baseline methods, validating the quality and consistency of explanations generated by IA-ViT. 

\section{Related Work}
\label{relatedworks}

\subsection{Explainable AI}
\label{sec:expmethod}
Depending on the method of explanation generation, general {\it post hoc} techniques in XAI can be broadly categorized into three groups: perturbation, approximation, and back-propagation. Perturbation methods, such as RISE \cite{petsiuk2018rise}, Extremal Perturbations \cite{fong2019understanding}, and SHAP \cite{lundberg2017unified}, attempt to generate explanations by purposely perturbing the input images. However, these methods are often characterized by time-consuming and inefficient performance in practical applications. Approximation methods employ an external agent as the explainer for black-box models, such as LIME \cite{ribeiro2016should} and FLINT \cite{parekh2021framework}. Nonetheless, these approaches might not accurately capture the true predictive mechanism of the models. Although back-propagation techniques apply the back-propagation scheme to generate gradient \cite{simonyan2013deep,zhou2016learning,selvaraju2017grad,li2023negative} or gradient-related \cite{bach2015pixel,shrikumar2017learning,sundararajan2017axiomatic,shrikumar2017learning,pan2021explaining,qiang2022attcat,li2023negative} explanations, these methods may not faithfully reveal the decision-making process of trained models and often demonstrate limited reliability and robustness \cite{adebayo2018sanity,kindermans2019reliability,agarwal2022openxai}. 

Different from {\it post hoc} methods, alternative methods suggest making alterations to either  architectures \cite{frosst2017distilling,wu2018beyond,al2020contextual,bohle2022b},  loss functions \cite{zhang2018interpretable,chen2020concept,ismail2021improving}, or both \cite{angelov2020towards,chen2019looks,pan2020explainable}. Nevertheless, certain methods depend on factors like the presence of ground truth explanations \cite{ghaeini2019saliency}, the accessibility of annotations concerning incorrect explanations for specific inputs \cite{frosst2017distilling}, or external knowledge sources \cite{al2020contextual}. Moreover, some interpretability constraints can potentially restrict the model's expressive capabilities, which may lead to a trade-off with prediction performance.

\vspace{-0.15cm}
\subsection{Explanation Methods for ViTs}

Motivated by the impressive success of Transformer architecture in NLP tasks  \cite{vaswani2017attention}, researchers have made efforts to extend the use of Transformer-based models to CV tasks \cite{dosovitskiy2020image,carion2020end,liu2021swin,touvron2021training,yuan2021tokens,zhou2021deepvit,chu2021twins,touvron2021going,wang2022bevt,wang2021pyramid,liu2022video,zhang2021vidtr,guo2022cmt}. Meanwhile, researchers have been actively exploring ways to enhance their interpretability. One popular approach involves analyzing the attention weights of MSA in ViTs \cite{vaswani2017attention,abnar2020quantifying}, however, the simple utilization may not provide reliable explanations \cite{serrano2019attention,qiang2022attcat}. Other approaches have been proposed to reason the decision-making process of ViTs, such as using gradients \cite{chen2022lctr,gao2021ts,gupta2022vitol,qiang2022attcat}, attributions \cite{chefer2021transformer,yuan2021explaining}, and redundancy reduction \cite{pan2021ia}.

Recently, some approaches have emerged to modify the ViT architecture to enhance interpretability. The Concept-Transformer \cite{rigotti2021attention}, for instance, exposes explanations of a ViT model's output in terms of attention over user-defined high-level concepts. However, the effectiveness of these methods heavily relies on the presence of these human-annotated concepts. \cite{kim2022vit} proposed ViT-NeT, which interprets the decision-making process through a tree structure and prototypes with visual explanations. Nevertheless, this method is not broadly applicable to various Transformer architectures and requires additional tree structures and external knowledge.

Differently, we propose IA-ViT to directly improve its interpretability during the training process with a novel interpretability-aware training objective. Moreover, our approach does not require external knowledge, such as pre-defined human-labeled concepts like Concept-Transformer \cite{rigotti2021attention} and additional complex architectures like ViT-NeT \cite{kim2022vit}. 

\section{Our Approach - IA-ViT}

\subsection{Problem Formulation}
In the context of explanation, conventional {\it post hoc} methods typically involve an \textit{explainer} module $\mathcal{G}$. This module takes the pre-trained model $\mathcal{F}$ and an input $x$ to produce an explanation $e$ for the output $y$, formally: $\mathcal{G}: \mathcal{F} \times x \to e$. The space of potential $e$ is usually determined by the specific explanation method in use. For example, a method employing saliency maps may define $e$ as normalized distributions indicating the importance of individual inputs, such as tokens and pixels.

In our work, we attempt to tackle a more general problem named \textit{Learning with Interpretation}, which advocates that the interpretation task should be integrated into the training process, as opposed to treating them as separate {\it post hoc} procedures. The core idea is to design a dedicated module, referred to as an interpreter, as an integral part of the model. This interpreter module relies on the predictor and is trained concurrently with it to provide interpretability for the trained model. Essentially, this approach augments the model's training process, encompassing not only the prediction objective but also an additional interpretability-aware objective.

Concretely, we propose a novel interpretability-aware training scheme to address the \textit{Learning with Interpretation} problem. Our training framework for IA-ViT consists of three key objectives for the minimization of dedicated losses and regularization terms as shown in Fig. \ref{fig:architecture}: (1) A primary objective focusing on target prediction, aiming to minimize Cross-Entropy loss $\mathcal{L}_\mathrm{ce}$; (2) An additional objective centered on simulation, which encourages the interpreter to emulate the behavior of the predictor, and this is quantified as $\mathcal{L}_\mathrm{kd}$ using knowledge distillation; (3) An attention regularizer that aligns the attention weights from the MSA blocks with the interpretable SSA block $\mathcal{L}_\mathrm{reg}$.  

\vspace{-0.05cm}
\subsection{IA-ViT Architecture}
The proposed IA-ViT framework consists of three components: feature extractor $h$, predictor $f$, and interpreter $g$, as shown in Fig. \ref{fig:architecture}. The feature extractor, comprising a stack of $L$ MSA blocks, takes the input image $x$ and encodes it into $\mathbf{z} \in \mathbb{R}^{(N + 1) \times d}$: $\mathbf{z} = h(x)$, where $N$ represents the number of image patches and $d$ is the embedding dimension. Subsequently, the predictor $f$ utilizes the feature embedding of the class token $\mathbf{z}^0$ from $\mathbf{z}$ to make predictions via a linear head: $\hat{y}_\mathrm{pred} = f(\mathbf{z}^0)$. Conversely, the interpreter $g$ takes the remaining feature embeddings as inputs, processing them through an SSA block followed by a linear head, to generate the prediction $\hat{y}_\mathrm{int} = g(\mathbf{z}^1,\cdots,\mathbf{z}^N)$. This linear head serves as the final classification layer of the interpreter, responsible for producing the ultimate predictions. It is a simple linear layer designed to preserve the interpretability of the interpreter and prevent the confusion of information that a more complex multi-layer perceptron (MLP) might cause. Additionally, the feature embeddings of the image patch tokens are not aggregated; instead, they are directly used as input to the linear head. Thus, IA-ViT employs both the predictor and the interpreter to generate two highly aligned predictions $\hat{y}_\mathrm{pred}$ and $\hat{y}_\mathrm{int}$, while sharing the feature extractor $h$.

\subsection{Interpretability of IA-ViT}

The rationale behind incorporating an interpreter into IA-ViT is to enhance its interpretability by gaining insights into its prediction process. It is crucial that the interpreter faithfully replicates the behavior of the predictor, ensuring that its output closely aligns with the predictor's output for a given input. Essentially, the predictor's role is to convey the crucial aspects of the input that influence the final prediction, while the interpreter complements this by offering supplementary insights into the model's decision-making process without altering the actual prediction.

Attention weights derived from MSA blocks can offer interpretable clues, but existing attention weights-based explanation methods \cite{serrano2019attention,abnar2020quantifying} only provide {\it post hoc} explanations, which are limited in their ability to provide faithful explanations of the model's decision-making process. To address this problem, the interpreter of IA-ViT applies an SSA mechanism, which dynamically aligns its attention weights with the discriminative patterns from the feature embeddings. This alignment offers more informative insights compared to the attention weights derived solely from the MSA blocks, which inherently combines the contributions of discriminative input patterns with respect to the model's outputs in an interpretable manner. It excels at emphasizing the specific input features that the model relied upon to make its predictions.

Given the input from the feature embeddings $\mathbf{Z}^\prime$, we obtain the projected key, query, and value as: 
\begin{equation}
    \mathbf{Q} = \mathbf{Z}^{\prime}\mathbf{W}^Q, \ \  \mathbf{K} = \mathbf{Z}^{\prime}\mathbf{W}^K,   \ \ \mathrm{and}  \ \ \mathbf{V} = \mathbf{Z}^{\prime}\mathbf{W}^V, 
\end{equation}
where $\mathbf{W}^Q \in \mathbb{R}^{d \times d}$, $\mathbf{W}^K \in \mathbb{R}^{d \times d}$, and $\mathbf{W}^V \in \mathbb{R}^{d \times d}$ are trainable transform matrices. Note $\mathbf{Z}^\prime$ does not contain the feature embedding of the class patch $\mathbf{z}^0$. 
Based on SSA Eq.\ref{eq:attweights},
\begin{equation}
    \label{eq:attweights}
    \mathbf{A} = \mathrm{Softmax}\bigg(\frac{\mathbf{Q}\mathbf{K}^T}{\sqrt{d_k}} \bigg) \ \ \mathrm{and} \ \ \mathbf{S} = \mathbf{A}\mathbf{V},
\end{equation}
we obtain the attention weights $\mathbf{A}$ that characterize the amount of attention paid to each patch and the SSA features $\mathbf{S}$. Then, we get $\|\mathbf{A}\| \leq 1$. Therefore, $\mathbf{S}$ is upper-bounded as: 
\begin{equation}
\label{eq:upperbound}
    \|\mathbf{S}\| = \|\mathbf{A}\|\ \|\mathbf{V}\|\cos(\mathbf{A},\mathbf{V}) \leq \|\mathbf{V}\|.
\end{equation}
When $\mathbf{S}$ is optimized, the attention weights $\mathbf{A}$ are proportional to $\mathbf{V}$. To achieve maximal output, $\mathbf{A}$ is driven to align with the discriminative features in $\mathbf{V}$. Consequently, $\mathbf{S}$ can only achieve this upper bound if all possible solutions of $\mathbf{v} \in \mathbf{V}$ are encoded as eigenvectors of $\mathbf{A}$. This maximization implies that, with the attention weights $\mathbf{A}$, we will obtain an inherently explainable decomposition of input patterns.

\subsection{Learning with Interpretation}
\label{sec:Learning with Interpretation}

Within the framework of \textit{Learning with Interpretation}, the interpreter's goal extends beyond optimizing predictions alone; it also involves comprehending the rationale behind the model's predictions concurrently. Therefore, IA-ViT adopts a joint training approach for the predictor and interpreter. This allows the interpreter to acquire insights that align with the predictions made by the predictor, ultimately enhancing the overall interpretability of the model. In this approach, the interpreter and predictor collaborate to produce accurate predictions while concurrently offering explanations for these predictions. This dual functionality can prove invaluable in various domains, including healthcare and finance, where the interpretability of learned models hold paramount importance.

\subsubsection{Classification Objective}
Given an input image $x$ with its corresponding label $y$, the final prediction is produced by the extractor and the predictor. Typically, the training process for the feature extractor and predictor involves minimizing the cross-entropy loss, which measures the disparity between the predicted probability distribution and the true labels. Formally, the cross-entropy loss is expressed as: 
\begin{equation}
\label{eq:celoss}
    \mathcal{L}_\mathrm{ce} = -\frac{1}{n}\sum_{i=1}^ny_i\log(f(h(x_i))),
\end{equation}
where $f$ and $h$ are the predictor and feature extractor components of IA-ViT, respectively.

\subsubsection{Simulation Objective} 

Knowledge distillation (KD) is a technique introduced in \cite{hinton2015distilling}, wherein a larger capacity teacher model is used to transfer its ``dark knowledge" to a more compact student model. The goal of KD is to achieve a student model that not only inherits better qualities from the teacher but is also more efficient for inference due to its compact size. A recent study \cite{alharbi2021learning} highlights the effectiveness of explainable knowledge distillation in transferring not only the performance but also the explanation information from the teacher model to the student model.

We use KD as a simulation objective in the \textit{Learning with Interpretation} framework. The simulation objective is formulated to force the interpreter's predictions to simulate the behavior of the predictor, as opposed to relying directly on ground truth labels but the soft labels generated by the predictor. In more detail, the logits generated by the predictor are denoted as $\mathbf{q} = [q_1, q_2, \cdots, q_C]$, which is the output distribution computed by applying softmax over the outputs: 
\begin{equation}
    \label{eq:logits}
    q_i = \frac{\exp(f(h(x))_i)}{\sum_{j=1}^C\exp(f(h(x))_j)},
\end{equation}
where $C$ is the number of classes. The logits are scaled by a temperature factor $\tau$ for a smooth distribution. Similarly, the interpreter produces a softened class probability distribution $\mathbf{p}$. Then KD is applied to the two probabilities: 
\begin{equation}
\label{eq:kdloss}
    \mathcal{L}_\mathrm{kd} =  -\frac{\tau^2 }{n} \sum_{i=1}^n (q_i/\tau)\log(p_i/\tau). 
\end{equation}

By optimizing $\mathcal{L}_\mathrm{kd}$, the interpreter is trained to predict the same class as the predictor with a high probability, enhancing the fidelity of interpretations to the model's outputs.

\subsubsection{Attention Regularization}
\label{sec:reg}
To further improve the interpretability of IA-ViT, we introduce an additional regularization term into the training objective. This term serves to reduce the Maximum Mean Discrepancy (MMD) \cite{gretton2006kernel,gretton2012kernel} between the attention distribution of MSA in the feature extractor, denoted as $\boldsymbol{\alpha}^\mathrm{E}$, and the attention distribution of the SSA in the interpreter, denoted as $\boldsymbol{\alpha}^\mathrm{I}$. This helps to ensure that the attention weights used by the feature extractor and the interpreter are generated from the same distribution, further improving the interpretability of the model.

Since MSA in the feature extractor employs multi-headed attention with multiple different attention vectors in each block, we aggregate these attentions by summing up the attention from the class token to other tokens in the last layer. This summation is then averaged across all attention heads to get $\boldsymbol{\alpha}^\mathrm{E}$. In contrast, $\boldsymbol{\alpha}^\mathrm{I}$ can be directly extracted from SSA in the interpreter. MMD compares the sample statistics between $\boldsymbol{\alpha}^\mathrm{I}$ and $\boldsymbol{\alpha}^\mathrm{E}$, and if the discrepancy is small, $\boldsymbol{\alpha}^\mathrm{I}$ and $\boldsymbol{\alpha}^\mathrm{E}$ are then likely to follow the same distribution. Thus, the attention regularizer is formulated as:
\begin{equation}
   \label{eq:attreg}
   \mathcal{L}_\mathrm{reg} = \mathrm{MMD}(\boldsymbol{\alpha}^\mathrm{I}, \boldsymbol{\alpha}^\mathrm{E}). 
\end{equation}

We conduct an in-depth analysis of this attention regularization to obtain a more comprehensive understanding of its positive impacts on the IA-ViT training process. Specifically, using the kernel trick, the empirical estimate of MMD $M$ can be obtained as:
\begin{equation}
\begin{aligned}
    \label{eq:mmd}
    M = \bigg[\frac{1}{n^2}\sum_{i,j = 1}^n \mathcal{K}(\alpha^{\mathrm{I}}_i,\alpha^{\mathrm{I}}_j) + \frac{1}{n^2}\sum_{i,j = 1}^n \mathcal{K}(\alpha^{\mathrm{E}}_i,\alpha^{\mathrm{E}}_j) \\
    - \frac{2}{n^2}\sum_{i = 1}^n \sum_{j = 1}^n \mathcal{K}(\alpha^{\mathrm{I}}_i,\alpha^{\mathrm{E}}_j)\bigg]^{1/2},
\end{aligned}
\end{equation}
where $\mathcal{K}(\cdot,\cdot)$ is a kernel function, and $n$ is the number of samples. Gretton et al. \cite{gretton2006kernel} showed if $\mathcal{K}$ is a characteristic kernel, then $\mathrm{MMD}(\boldsymbol{\alpha}^\mathrm{E}, \boldsymbol{\alpha}^\mathrm{I})$ = 0 asymptotically if and only $\boldsymbol{\alpha}^{\mathrm{I}}$ and $\boldsymbol{\alpha}^{\mathrm{E}}$ are from the same distribution. A typical choice of $\mathcal{K}$ is the Gaussian kernel with bandwidth parameter $\sigma$: 
\begin{equation}
    \mathcal{K}(x, y) = \exp\bigg(\frac{-\|x - y\|^2}{\sigma}\bigg).
\end{equation}
With the Gaussian kernel, minimizing MMD is equivalent to matching all orders of moments of the two distributions.

Inspired by the idea of \cite{li2023learning}, we further analyze the effect of MMD on our regularization. Since $\boldsymbol{\alpha}^{\mathrm{I}}$ and $\boldsymbol{\alpha}^{\mathrm{E}}$ are symmetric in MMD, we only present the attention weights of $\boldsymbol{\alpha}^{\mathrm{I}}$ here without loss of generality. We first formulate the gradient of the regularization loss with respect to $\boldsymbol{\alpha}^{\mathrm{I}}$ as: 
\begin{equation}
    \nabla_{\alpha_i^{\mathrm{I}}} M = \frac{2}{\sqrt{M}} \nabla_{\alpha^{\mathrm{I}}}\bigg[\frac{1}{n^2}\sum_{j = 1}^n \mathcal{K}(\alpha^{\mathrm{I}}_i, \alpha^{\mathrm{I}}_j) 
    - \frac{2}{n^2}\sum_{j = 1}^n \mathcal{K}(\alpha^{\mathrm{I}}_i,\alpha^{\mathrm{E}}_j) \bigg]. 
\end{equation}

The gradient with respect to $x$ for Gaussian kernel $\mathcal{K}$ is:
\begin{equation}
    \label{eq:kgradient}
    \nabla_x\mathcal{K}(x,y)= -2\exp \bigg(\frac{-\|x-y\|^2}{\sigma} \bigg) \cdot \frac{x-y}{\sigma}. 
\end{equation}
$\sigma$ here is a data-dependent hyperparameter and not back-propagated in the training process. We thus get
\begin{equation}
\begin{aligned}
    \nabla_{\alpha_i^{\mathrm{I}}}M & = -\frac{2}{\sqrt{M}}\bigg[\frac{1}{n^2}\sum_{j=1}^n\exp\bigg(-\frac{\|\alpha^{\mathrm{I}}_i - \alpha^{\mathrm{I}}_j\|^2}{\sigma}\bigg)\cdot \frac{\alpha^{\mathrm{I}}_i - \alpha^{\mathrm{I}}_j}{\sigma} \\
    & \qquad -\frac{2}{n^2}\sum_{j=1}^n\exp\bigg(-\frac{\|\alpha^{\mathrm{I}}_i - \alpha^{\mathrm{E}}_j\|^2}{\sigma}\bigg)\cdot \frac{\alpha^{\mathrm{I}}_i - \alpha^{\mathrm{E}}_j}{\sigma} \bigg],
\end{aligned}
\end{equation}
by the linearity of the gradient operator. We notice that for function $g_a(x) = \exp({-x^2/a})x/a$ ($a$ is a constant), $g_a(x) \to 0$ exponentially as $x \to \infty$. We further achieve 
\begin{equation}
\begin{aligned}
    \label{eq:ineq} 
    \|\nabla_{\alpha^{\mathrm{I}}}M\| \leq \frac{2}{\sqrt{M}} \left[\frac{1}{n^2}\sum_{j=1}^n g_\sigma(\|\alpha^{\mathrm{I}}_i - \alpha^{\mathrm{I}}_j\|) \right. 
        \\ 
    + \left. \frac{2}{n^2}\sum_{j=1}^n g_\sigma(\|\alpha^{\mathrm{I}}_i - \alpha^{\mathrm{E}}_j\|) \right]
\end{aligned}
\end{equation}
using the triangle inequality for fixed $\sigma$. $\sqrt{M}$ here is a constant for all samples within the training mini-batch.

We observe that when $\alpha^{\mathrm{I}}$ deviates significantly away from the majority of samples of the same class, i.e., noisy samples or outliers, $\|\alpha^{\mathrm{I}}_i - \alpha^{\mathrm{I}}_j\|$ and $\| \alpha^{\mathrm{I}}_i - \alpha^{\mathrm{E}}_j \|$ are large, the magnitude of its gradient in the regularization loss diminishes from Eq.\ref{eq:ineq}. More specifically, $\alpha^{\mathrm{I}}$ has negligible impact on the regularization term. On the other hand, training IA-ViT with the regularization term promotes the alignment of attention weights representations of samples that stay close in attention weights distribution. The attention weights deviating from the majority are likely low-density or even outliers from the distribution perspective. Overall, such behavior of the regularization loss implies that it can help IA-ViT better capture information from high-density areas and reduce the distraction of low-density areas in learning feature representations on the data manifold, as shown in Fig \ref{fig:examples}.

\subsubsection{Overall Objective}
The overall training objective is formulated as the weighted sum of $\mathcal{L}_\mathrm{ce}$, $\mathcal{L}_\mathrm{kd}$, and $\mathcal{L}_\mathrm{reg}$. Formally, it is expressed as:
\begin{equation}
    \label{eq:overall}
    \mathcal{L} = \beta \cdot \mathcal{L}_\mathrm{ce} + (1 - \beta) \cdot(\mathcal{L}_\mathrm{kd} + \mathcal{L}_\mathrm{reg}),
\end{equation}
where $\beta \in (0,1)$ is a hyperparameter that balances the contributions of each term. 

\section{Experiment Settings}

\subsection{Model Architectures}
We employ the vanilla ViT-B/16 architecture \cite{dosovitskiy2020image} as the transformer backbone for our model. Specifically, we use the base version with patches of size $16 \times 16$, which was exclusively pre-trained on the ImageNet-21k dataset. This backbone consists of 12 stacked MSA blocks, each containing 12 attention heads. The model utilizes a total of 196 patches, and each patch is flattened and projected into a 768-dimensional vector. Positional embeddings are added to these patch embeddings, and the resulting embeddings are then processed by the feature extractor. Following this, the predictor utilizes the feature embeddings of the class patch and passes them through two fully connected layers and a softmax layer to produce logits for prediction. In contrast, the interpreter operates on the feature embeddings from other image patches. It employs a single SSA block, followed by two fully connected layers and a softmax layer, to generate logit scores for interpretation.

\vspace{-0.2cm}
\subsection{Baseline Explanation Methods}

\textbf{RawAtt} \cite{vaswani2017attention} leverages the attention weights from the first block of ViT to identify the most important patches for predictions. \textbf{Rollout} \cite{abnar2020quantifying} is another attention weights based explanation approach, which produces an explanation taking into account all the attention weights computed along the forward pass. \textbf{AttGrads} \cite{barkan2021grad} utilizes the gradients of the attention weights to pinpoint the most significant patches. \textbf{AGCAM} \cite{leem2024attention} as an attention-guided visualization method is proposed to leverage aggregated gradients which are guided by attention weights demonstrating to generate more faithful explanations. 

\vspace{-0.2cm}
\subsection{Evaluation Metrics}

To evaluate the IA-ViT model's performance comprehensively, we report accuracy metrics for both the predictor and the interpreter. We employ attribution maps, which are visual representations highlighting the input pixels considered significant or insignificant in relation to a predicted label. This approach is used for a qualitative evaluation of the explanation quality. Furthermore, we utilize insertion score and deletion score as quantitative evaluation metrics. In the first round of experiments, we replace the most important pixels with black pixels, following the approach of \cite{petsiuk2018rise}. In the second round, we replace these pixels with Gaussian-blurred pixels, as per \cite{sturmfels2020visualizing}. We report the average performance across both rounds of experiments. Since both deletion and insertion scores can be influenced by shifts in distribution when pixels are removed or added, we employ the difference between the insertion and deletion scores as an additional metric for comparison \cite{shah2021input}. Focusing on their relative differences helps mitigate the impact of the distribution shifts.

\section{Results and Discussion}

\subsection{Model Performance Evaluations}

Table \ref{accres} presents a performance comparison between IA-ViT and the vanilla ViT models. Both the ViT models and the predictor in IA-ViT achieve promising performance on these image classification tasks. The interpreter in IA-ViT also achieves performance {\it on par} with the predictor, largely owing to the adoption of the simulation objective. It is important to highlight that the IA-ViT models' final predictions rely on the predictor's outputs, as shown in Fig. \ref{fig:architecture}. We further use Performance Drop Rate (PDR) to evaluate the performance degradation, formally: $\mathrm{PDR} = 1 - \frac{\mathrm{Accuracy}_\mathrm{IA-ViT}}{\mathrm{Accuracy}_\mathrm{ViT}}$. 
The average PDR among these datasets is 1.16\%, indicating a non-substantial decrease in accuracy when employing the IA-ViT model with its integrated interpreter.

\begin{table}[t]
\scriptsize
\centering
\caption{Comparison of the classification accuracies of the ViT and IA-ViT. PDR refers to performance drop rate.}\label{accres}
\vspace{-0.1cm}
\begin{tabular}{c|c|c|ccc}
\hline
\multirow{2}{*}{Datasets} & \multirow{2}{*}{ViT} & \multicolumn{3}{c}{IA-ViT}                \\ 
                          &                     & \multicolumn{1}{c}{Predictor} & Interpreter & PDR (\%) \\ \hline
CIFAR10                   & 98.93                & \multicolumn{1}{c}{97.51}     & 97.24 &  1.43   \\
STL10                     & 99.31                & \multicolumn{1}{c}{97.73}     & 95.42 &  1.59    \\ 
Dog\&Cat                  & 99.72                & \multicolumn{1}{c}{98.82}     & 97.76 &  0.90    \\
CelebA                    & 96.87                & \multicolumn{1}{c}{96.16}     & 96.09 &  0.73    \\ \hline
\end{tabular}
\vspace{-0.1cm}
\end{table} 

\begin{figure*}[t]
    \centering 
    \includegraphics[width=0.98\linewidth]{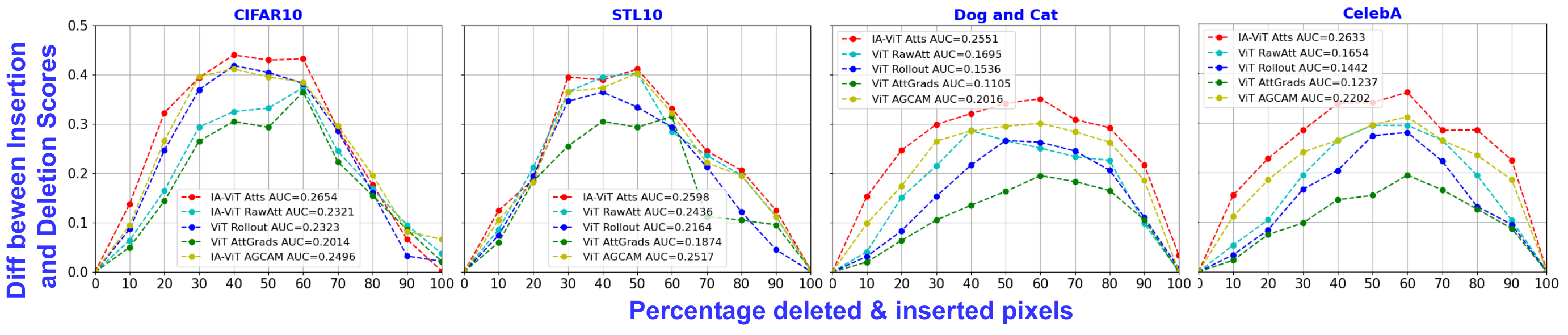}
    \vspace{-0.2cm}
    \caption{Quantitative explanation performance comparison in terms of differences between insertion and deletion scores.}
    \label{fig:quant}
    \vspace{-0.2cm}
\end{figure*}

\vspace{-0.1cm}
\subsection{Quantitative Explanation Evaluations}

\begin{table}
\scriptsize
\centering
\caption{Quantitative explanation performance comparison using deletion (D$\downarrow$) and insertion (I$\uparrow$) scores. The deletion score is the lower the better, while the insertion score is the higher the better. The best results are in bold.}\label{tab:quant}
\begin{tabular}{c|c|cccc|c}
\hline
\multirow{2}{*}{Datasets} & \multirow{2}{*}{M} & \multicolumn{4}{c|}{ViT}                & IA-ViT \\ 
                          &                          & \multicolumn{1}{l}{RawAtt} & Rollout & AttGrads & AGCAM & Atts     \\ \hline
\multirow{2}{*}{CIFAR10}  & D        & \multicolumn{1}{l}{0.3714} & 0.3817       & 0.3036  & 0.2841         & \textbf{0.2479}         \\ 
                          & I          & \multicolumn{1}{l}{0.6237}        & 0.6141 &  0.5583  &  0.6533      &  \textbf{0.7082}        \\ \hline
\multirow{2}{*}{STL10}    & D          & \multicolumn{1}{l}{0.3921}   & 0.3874      & 0.4124  & 0.3659          &  \textbf{0.3254}        \\ 
                          & I          & \multicolumn{1}{l}{0.5862} & 0.5967       &  0.5546  &  0.6124      &  \textbf{0.6436}        \\ \hline
\multirow{2}{*}{Dog\&Cat}              & D          & \multicolumn{1}{l}{0.6649} & 0.6785       & 0.7354 & \textbf{0.6158}        & 0.6232         \\ 
                          & I         & \multicolumn{1}{l}{0.8376} & 0.8322      & 0.7921   & 0.8741       & \textbf{0.8783}         \\ \hline
\multirow{2}{*}{CelebA}   & D         & \multicolumn{1}{l}{0.7131} & 0.7260       & 0.7536  & 0.6514        &  \textbf{0.5977}        \\ 
                          & I          & \multicolumn{1}{l}{0.8166}  & 0.8275       & 0.8123  & 0.8367        &  \textbf{0.8719}        \\ \hline
\end{tabular}
\vspace{-0.4cm}
\end{table} 

The quantitative evaluations shown in Table \ref{tab:quant} demonstrate that directly leveraging the attention weights (Atts) from the interpreter in IA-ViT as explanations outperform the baselines, i.e., RawATT, Rollout, AttGrads, and AGCAM, for ViT in terms of deletion and insertion scores across all datasets. The last column, representing as Atts in IA-ViT, achieves the smallest deletion scores and the largest insertion scores across most datasets. This further illustrates the explanations generated by the interpreter of IA-ViT effectively capture the most important discriminative pixels or patches for the image classification tasks. Similarly, the results of the difference between insertion and deletion scores across a varying percentage of deleted/inserted pixels, as shown in Fig. \ref{fig:quant}, demonstrate that the form of the interpreter in IA-ViT outperforms the other baselines in terms of Area Under the Curve (AUC) among all tasks. These quantitative evaluations collectively provide compelling evidence of IA-ViT's superior interpretability compared to the {\it post hoc} methods designed for ViT.

\begin{figure}[t]
    \centering 
    \includegraphics[width=0.90\linewidth]{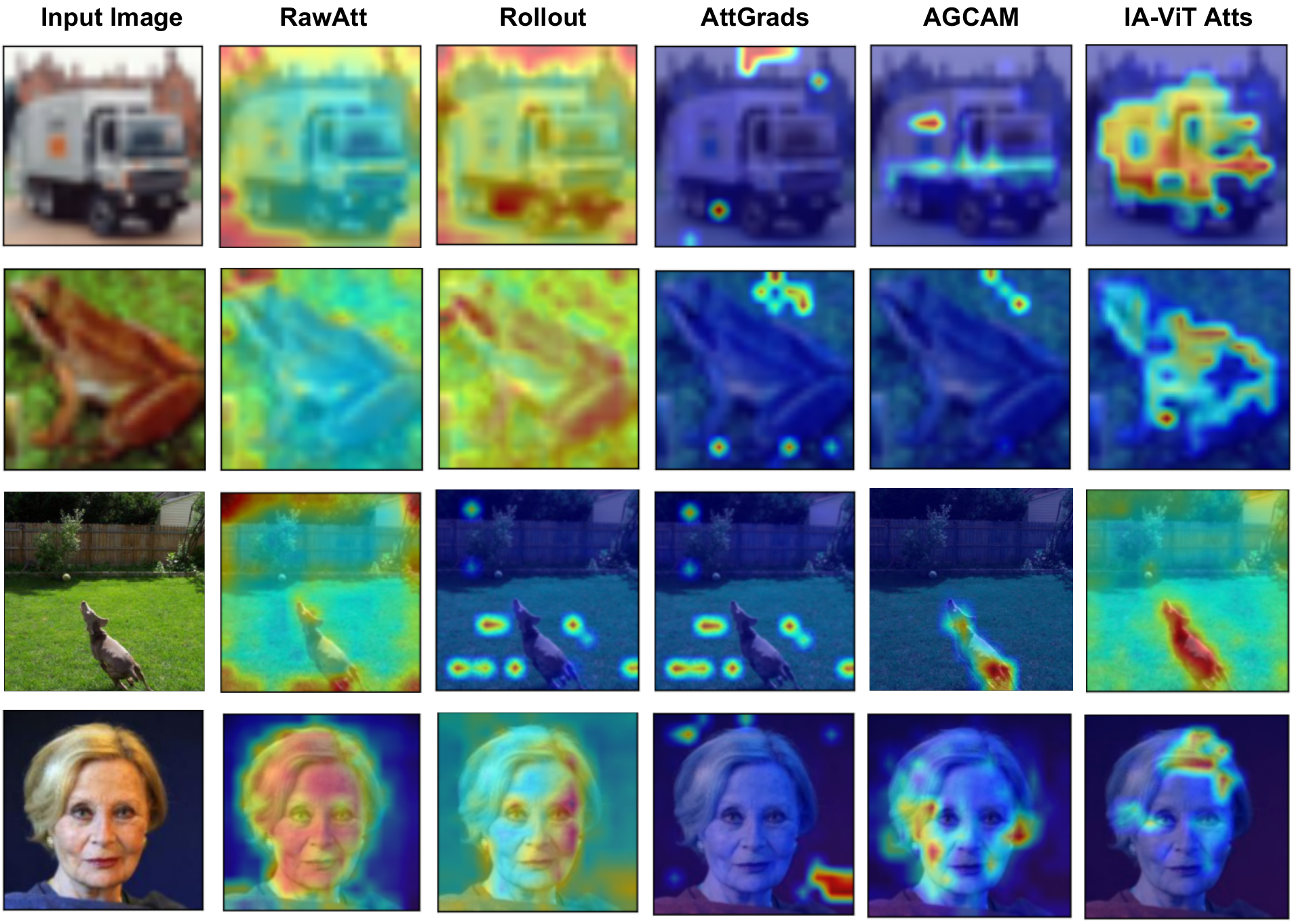}
    \vspace{-0.15cm}
    \caption{Examples of attribution maps generated by the baseline explanation methods for ViT and the attention weights from the interpreter of IA-ViT.}
    \label{fig:examples}
    \vspace{-0.85cm}
\end{figure}

\vspace{-0.1cm}
\subsection{Qualitative Explanation Evaluations}

The examples provided in Fig.\ref{fig:examples} vividly illustrate the superior quality of the attribution maps produced by IA-ViT's interpreter when compared to the {\it post hoc} baseline explanation methods, i.e., RawATT, Rollout, AttGrads, and AGCAM, designed for ViT. A key observation from this figure is that the heatmaps generated by IA-ViT's interpreter exhibit more focused attention on the target objects, whereas the heatmaps generated by other baselines, such as RawAtt and Rollout, are dispersed across both the background and class entities. In contrast, AttGrads produces heatmaps that primarily highlight areas unrelated to the target. While the most recent work AGCAM is also able to capture a small portion of the important regions of the targeted objects, e.g., the first and third rows, the IA-ViT's interpreter can capture the most important patches for the target objects prediction. It is essential to emphasize that the results depicted in Fig.\ref{fig:examples} are randomly selected from the four benchmark datasets, which are representative of the typical outcomes observed in our experiments.

Additionally, these qualitative explanation examples highlight the effectiveness of the attention regularization utilized in the training objective. IA-ViT models possess the capability to extract information from regions with high information density while mitigating the influence of regions with low information density during the feature learning process. Therefore, the interpreter produces high-quality explanations that densely emphasize the target object. This is clearly evident in Fig. \ref{fig:examples}, where the heatmaps generated by the interpreter distinctly highlight the target objects (e.g., hair, dog, truck, and frog) while disregarding the background or other irrelevant noise. In contrast, the explanations generated by the baselines merely accentuate certain irrelevant areas and fail to capture the precise shape of the target objects.

\vspace{-0.3cm}
\subsection{Ablation Study}
\vspace{-0.2cm}

The results shown in Table \ref{tab:ablation} highlight the importance of each component in the training objective as defined in Eq.\ref{eq:overall}. These ablation study results reveal that removing the simulation objective, i.e., without using $\mathcal{L}_\mathrm{kd}$ (Eq.\ref{eq:kdloss}), leads to significant drops in the accuracies of the interpreter. This decline is attributed to the interpreter's inability to mimic the predictor's prediction behavior in the absence of the simulation objective. Consequently, the interpreter fails to deliver accurate predictions for image classification tasks, defaulting to random guesses in binary classification tasks, such as Dog\&Cat and CelebA, where accuracies are around 0.5. Meanwhile, the interpreter is unable to generate meaningful explanations, resulting in larger deletion scores and smaller insertion scores as shown in Table \ref{tab:ablation}. Additionally, while removing the regularization term, i.e., without using $\mathcal{L}_\mathrm{reg}$ (Eq. \ref{eq:mmd}) from the training objective does not visibly affect the predictor's and interpreter's classification abilities per the Table, its absence markedly affects the quality of explanations. Specifically, it leads to a substantial increase in deletion scores and a decrease in insertion scores for the explanations compared to when the full training objective. In summary, incorporating all three terms in the training objective leads to both effective prediction performance and meaningful explanations.

\begin{table}[t]
\scriptsize
\centering
\caption{Ablation study of training losses of IA-ViT on Dog\&Cat and CelebA datasets. w/o denotes without.}\label{tab:ablation}
\vspace{-0.2cm}
\begin{tabular}{c|c|cccc}
\hline
\multirow{2}{*}{Datasets} & \multirow{2}{*}{Losses} & \multicolumn{4}{c}{IA-ViT}                \\ 
                          &                     & Predictor & Interpreter & Deletion  & Insertion  \\ \hline
\multirow{3}{*}{Dog\&Cat} & all                & 98.82     & 97.76 & 0.6232 & 0.8783    \\
                          & w/o $\mathcal{L}_\mathrm{kd}$               &  99.04    & 49.40 & 0.7085 & 0.8374   \\
                          & w/o $\mathcal{L}_\mathrm{reg}$               & 99.60     & 99.62 & 0.7264 & 0.8397   \\ \hline
\multirow{3}{*}{CelebA}   & all                & 96.16     & 96.09 & 0.5977 & 0.8719  \\
                          & w/o $\mathcal{L}_\mathrm{kd}$                & 95.65     & 49.69 & 0.6642 & 0.8134    \\
                          & w/o $\mathcal{L}_\mathrm{reg}$               & 97.07     & 97.04 & 0.6672 & 0.8216    \\ \hline
\end{tabular}
\vspace{-0.35cm}
\end{table} 

\vspace{-0.2cm}
\subsection{Fairness Learning} 
\vspace{-0.2cm}
\label{sec:fariness}

The examples from CelebA, specifically the hair color prediction task, illustrate that the attribution maps produced by the interpreter of IA-ViT concentrate intensely on the hair region, prioritizing it over other facial features. On the contrary, the explanations generated by the Rollout method demonstrate that vanilla ViT tends to learn spurious features that might be related to the sensitive attribute (in this case, gender) but not the real feature that is relevant to the hair color prediction. Table \ref{fariness} demonstrates that the IA-ViT model outperforms the ViT model in both fairness metrics on the hair color prediction task in the CelebA dataset. The reduced demographic parity (DP) and equality of odds (EO) values indicate that IA-ViT's training effectively mitigates bias, resulting in a fairer model. This further demonstrates the effectiveness of our interpretability-aware training, which indeed extracts ``real" features rather than spurious ones. '
\begin{table}
\scriptsize
\centering
\caption{Fairness and accuracy comparison for ViT and IA-ViT over the target (Y) and sensitive (S) on CelebA.}\label{fariness}
\vspace{-0.1cm}
\begin{tabular}{c|c|c|c}
\hline
\multirow{2}{*}{Models} & \multicolumn{3}{c}{$Y$: Hair Color\ \ $S$: Gender}   \\ \cline{2-4} 
& ACC$\uparrow$ & DP$\downarrow$ & EO$\downarrow$ \\ 
\hline
ViT      & 96.89 & 12.95 & 8.69  \\ 
IA-ViT & 96.59 & 9.81  & 5.76  \\ \hline
\end{tabular}
\vspace{-0.5cm}
\end{table} 

\vspace{-0.15cm}
\section{Conclusion}
\vspace{-0.15cm}

In this work, we propose an interpretability-aware variant of ViT named IA-ViT. Our motivation stems from the consistent predictive distributions and attention maps generated by both the CLS and image patches. IA-ViT consists of three major components: a feature extractor, a predictor, and an interpreter. By training the predictor and interpreter jointly, we enable the interpreter to acquire explanations that align with the predictor's predictions, enhancing the overall interpretability. As a result, IA-ViT not only maintains strong predictive performance but also delivers consistent, reliable, and high-quality explanations. Extensive experiments validate the efficacy of our interpretability-aware training approach in improving interpretability across various benchmark datasets when compared to several baseline explanation methods.

\bibliographystyle{IEEEtran}
\bibliography{references}

\end{document}